\documentclass{article}

\PassOptionsToPackage{numbers, compress}{natbib}

\usepackage[preprint]{neurips_2026}

\usepackage[utf8]{inputenc}
\usepackage[T1]{fontenc}
\usepackage{xurl}
\usepackage{hyperref}
\hypersetup{hidelinks}
\usepackage{url}
\usepackage{booktabs}
\usepackage{amsfonts}
\usepackage{amssymb}
\usepackage{amsmath}
\usepackage{amsthm}
\usepackage{mathtools}
\usepackage{nicefrac}
\usepackage{xcolor}
\usepackage{graphicx}
\usepackage{enumitem}
\usepackage{algorithm}
\usepackage{algpseudocode}

\newcommand{\ourMethod}{\textsc{SkillRAE}}

\title{SkillRAE: Agent Skill-Based Context Compilation for Retrieval-Augmented Execution}

\author{%
  Xiangcheng Meng ~~~~ Shu Wang ~~~~ Yixiang Fang\thanks{Corresponding author.}\\
  The Chinese University of Hong Kong, Shenzhen\\
  \texttt{xiangchengmeng@link.cuhk.edu.cn, shuwang3@link.cuhk.edu.cn, fangyixiang@cuhk.edu.cn}\\
}

\begin{document}

\maketitle

\begin{abstract}
Large Language Model (LLM)-based agents (e.g., OpenClaw) increasingly rely on reusable skill libraries to solve artifact-rich tasks such as document-centric workflows and data-intensive analysis. As these libraries grow, a few works have attempted to study the Retrieval-Augmented Execution (RAE), which often first retrieves some external skills and other knowledge, then compiles the context using retrieved skills, and finally executes the task. Existing works mainly focus on optimizing skill retrieval and task execution, and they pay little attention to how to effectively organize the selected skill evidence in a form that is compact, grounded, and immediately usable for the downstream executors to complete tasks. To fill this gap, we propose \ourMethod{}, a two-stage RAE approach focusing on skill-based context compilation, which consists of the offline and online stages. Specifically, in the offline indexing stage, it builds a multi-level skill graph over skill communities, skills, and reusable subunits, for capturing their relationships. In the online retrieval stage, it first performs skill-ranked retrieval with selected-subunit evidence export in the graph, and then applies rescue-aware compact compilation to recover the key evidence. Together, these components compile a coarse-ranked skill set into a task-specific context that is compact, grounded, and immediately usable. Experiments on two public benchmarks show that \ourMethod{} achieves a significant improvement over baselines for RAE. For example, on SkillsBench, it achieves an improvement of \textbf{11.7\%} over the SOTA method. Ablation studies further show that our context compilation is crucial, instead of a mere prompt addition.

\end{abstract}

\section{Introduction}
\label{sec:intro}
\begin{figure}[t]
    \centering
    \includegraphics[width=0.95\linewidth]{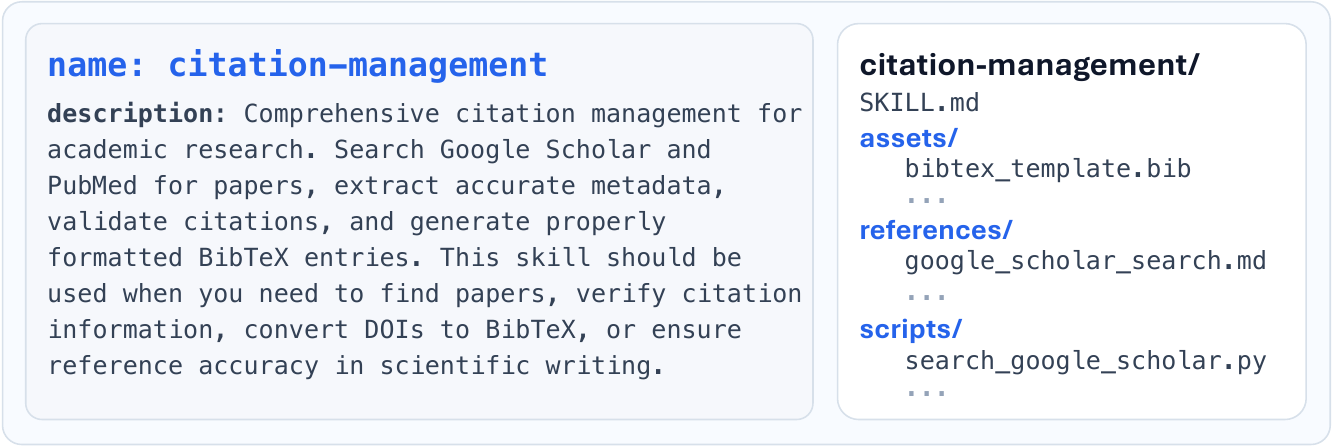}
    \caption{An example agent skill from SkillsBench.
    It contains a natural-language description, together with optional linked files such as assets, references, and scripts.}
    \label{fig:skill-example}
\end{figure}

LLM-based agents, such as OpenClaw and Manus, are increasingly deployed as capability-access systems to complete artifact-rich tasks such as document-centric workflows and data-intensive analysis \citep{openclaw2026,manus2026docs,li2023apibank,qin2024toolllm,ye2025toolhop,shen2025shortcutsbench}.
Instead of relying only on parametric model knowledge, they often use pre-defined agent skills to execute tasks.
Conceptually, an agent skill is a reusable procedural artifact that encode the specific ``how-to'' knowledge for coordinating tools, memory, and runtime context under concrete constraints \citep{wang2023voyager,xli2026skillsbench,hli2026agentskillos}.
When there is no ambiguity, we simply call it a skill.
Figure~\ref{fig:skill-example} shows a real \texttt{citation-management} skill from SkillsBench, illustrating that an agent skill is a reusable bundle rather than a single tool name.
As shown in recent works, many skill libraries, such as SkillBench \citep{xli2026skillsbench} and AgentSkillOS \citep{hli2026agentskillos}, have been constructed and released.

Recently, a few works have attempted to study the {\it Retrieval-Augmented Execution (RAE)}, which often first retrieves some external skills and other knowledge, then compiles the context using retrieved skills, and finally executes the task.
%
These works can roughly be classified into three categories:
1) Skill routing methods focus on identifying relevant APIs or skills from large candidate pools \citep{li2023apibank,qin2024toolllm,ye2025toolhop,shen2025shortcutsbench,zheng2026skillrouter}.
2) Repository-aware retrieval methods show that code, document, and graph structure can improve evidence access beyond flat vector search \citep{zhang2023repocoder,cheng2024dataflow,ouyang2025repograph,sarthi2024raptor,gutierrez2024hipporag,li2025structrag,wang2026archrag,wang2025bookrag}.
3) Execution planning methods further decompose tasks and assemble multi-step tool-use trajectories or execution plans \citep{shen2023hugginggpt,wu2024toolplanner}.
However, these works mainly focus on optimizing skill retrieval and task execution, and they pay little attention to skill-based context compilation and face a key issue: {\it A retrieved skill set may be globally relevant, but still under-resolved for execution.}
This is because skill routing methods usually stop at ranked skills, repository-aware retrieval methods improve evidence access without compiling a skill-facing context artifact, and execution planning methods often change the execution surface by introducing explicit plans or control policies.
Hence, what remains missing is a compilation mechanism that can effectively convert the selected skill evidence in a form that is compact, grounded, and immediately usable for the downstream executors to complete tasks.

This gap is nontrivial for two reasons.
First, individual skills are often designed for customizing some specific tasks. While many real-world tasks are highly relevant to these skills, they cannot use them directly, since the subunits (e.g., short procedure, file convention, usage note, etc.) of a skill may be critical for solving many tasks, leading to high relevance.
%
%
Second, isolated subunits of skills cannot be used for solving a task either, though they may be relevant to it, because the subunits of a skill often have dependence relationship with some constraints \citep{cheng2024dataflow,ouyang2025repograph,wang2026archrag,wang2025bookrag}.
Here, a \textit{subunit} refers to a small yet reusable unit extracted from a skill, such as a script, an execution step, or a procedural instruction.
For example, the ``generate properly formatted BibTeX entries" in Figure~\ref{fig:skill-example} can be regarded as a subunit.
%
%
%
Hence, the key challenge is not just select which individual skills or subunits are relevant, but how to effectively compile the relevant skills in a form that is compact, grounded, and immediately usable. 

To fill the above gap, we propose \ourMethod{}, a two-stage RAE approach for compiling the skill-based context; that is, given a task request and a repository of reusable skills, we aim to identify a set of relevant skills and subunits, and then compile them into a task-specific context that is compact, grounded, and immediately usable.
Specifically, \ourMethod{} consists of an offline stage and an online stage.
In the stage, it builds a multi-level skill graph that includes skill nodes, subunit nodes, skill descriptions, skill communities, and their link relationships.
Serving as a powerful repository, this graph can well capture the relationships between these different nodes.
%
In the online stage, \ourMethod{} first retrieves some relevant skills.
Next, it identifies a small number of highly relevant subunits from other skills that are not selected.
Afterwards, to make the context compact, grounded, and directly reusable for agents, it synthesizes the identified subunits into the most relevant retrieved skills.
Finally, it serializes the retrieved skills into a task-specific context with some task-specified guidance on how to apply these skills effectively.
%
%

We have conducted extensive experiments on two public benchmarks, and the results show that \ourMethod{} significantly outperforms the SOTA baselines.
For example, on SkillsBench with 87 tasks, \ourMethod{} achieves a mean verifier reward of 29.26\%, improving over pure curated skills by 11.7\% relative and over the strongest automated baseline by 32.8\% relative.
Ablation studies further show that context compilation is crucial for forming an executable skill context, rather than serving as a simple prompt addition.

In summary, our main contributions are summarized as follows:
\begin{itemize}[nosep]


    
    \item We propose \ourMethod{}, a two-stage RAE approach that builds a multi-level skill graph in the offline stage, and retrieves skills and compiles the context in the online stage.
    
    \item We propose a novel context compilation method, which compiles both the highly relevant skills and subunits of skills and  generates some task-specified guidance.
    
    \item We extensively evaluate \ourMethod{} from different angles through downstream agent benchmarks, and the results show that \ourMethod{} achieves superior performance.
\end{itemize}

\section{Related Work}
\label{sec:related}

\textbf{Agent Skills.}
A skill is a reusable procedural artifact with bounded scope that externalizes task-focused know-how, including when to act, how to execute, how to judge completion, and how to expose instructions progressively at execution time~\cite{ni2026trace2skill,ouyang2025reasoningbank}.
This makes skills different from APIs or tool-use interfaces, which mainly specify callable operations rather than reusable execution guidance.
Recent works have begun to construct skill libraries, such as SkillsBench~\cite{xli2026skillsbench} which provides an execution-centric benchmark with deterministic verifiers and AgentSkillOS~\cite{hli2026agentskillos} which organizes skills at ecosystem scale under structured orchestration pipelines.
Besides benchmark construction, some works have studied how agents accumulate reusable skills and how dependency-aware repositories can support skill combination and task routing~\cite{liu2026graphofskills,zheng2026skillrouter}.

\textbf{Tool and API Selection.}
Tool use and API calls extend agents by giving them callable external operations for information access, computation, and environment interaction.
Tool-use methods establish prompting and training paradigms for deciding when to call tools \citep{yao2023react,schick2023toolformer}.
API-call works further scale this process by indexing tool names, descriptions, schemas, documentation, demonstrations, and ranking candidate APIs before downstream invocation \citep{li2023apibank,qin2024toolllm,patil2024gorilla,hao2023toolkengpt,du2024anytool,chen2024reinvoke}.
These works focus on studying how to select and invoke callable operations, whereas {\ourMethod} addresses how skill evidence can be compiled into a compact, grounded, and immediately usable form.

\textbf{Retrieval-Augmented Generation (RAG).}
RAG grounds LLMs by retrieving external evidence and conditioning generation on the context \citep{Lewis2020RAG}.
Existing RAG methods can be classified into three categories: vector database-based method \cite{zhou2025depth}, graph-based methods \cite{wang2026archrag,wang2025bookrag,gutierrez2024hipporag,sarthi2024raptor,edge2024graphrag,chen2026pathrag,guo2024lightrag,huang2025retrieval}, and agentic methods which retrieve in an iterative action~\cite{jiang2023active,asai2023self,trivedi2023interleaving,yang2025demystifying}.
Note that RAG methods often have an additional post-processing step which includes passage filtering, summarization, compression, and rewriting before LLM-context placement~\cite{xu2023recomp,jiang2024longllmlingua,fang2024trace}.
For example, RECOMP~\cite{xu2023recomp} and LongLLMLingua~\cite{jiang2024longllmlingua} compress retrieved context, while TRACE organizes evidence into reasoning chains.
Such methods retrieve factual knowledge, whereas {\ourMethod} retrieves and compiles skills which contain not only knowledge, but also tool use and API calls.

\textbf{Retrieval-Augmented Execution (RAE).}
RAE extends the retrieval-augmentation paradigm from knowledge-grounded generation to agent execution, where external procedural artifacts, such as skills, tools, APIs, and so on, are retrieved to execute a specific task.
Existing RAE works can be grouped into three categories:
1) Skill routing methods focus on selecting, structuring, or composing reusable skills from large skill ecosystems~\cite{zheng2026skillrouter,liu2026graphofskills,hli2026agentskillos}.
2) Repository-aware retrieval methods exploit code, dataflow, document, or graph structure to improve access to execution-relevant evidence beyond flat vector search~\cite{zhang2023repocoder,cheng2024dataflow,ouyang2025repograph}.
3) Execution planning methods further connect retrieval with explicit execution planning or orchestration by assembling multi-step tool-use trajectories, model calls, or skill pipelines~\cite{shen2023hugginggpt,lu2023chameleon,wu2024toolplanner,hli2026agentskillos}.
In contrast, \ourMethod{} focuses on the compilation step between retrieval and execution, converting relevant skills and subunits into a compact, grounded, and immediately usable context, without changing the downstream executor.

\section{Our RAE Approach: \ourMethod}
\label{sec:method}
\subsection{Overview}
\label{sec:method-overview}

\begin{figure}[t]
    \centering
    \includegraphics[width=\textwidth,clip]{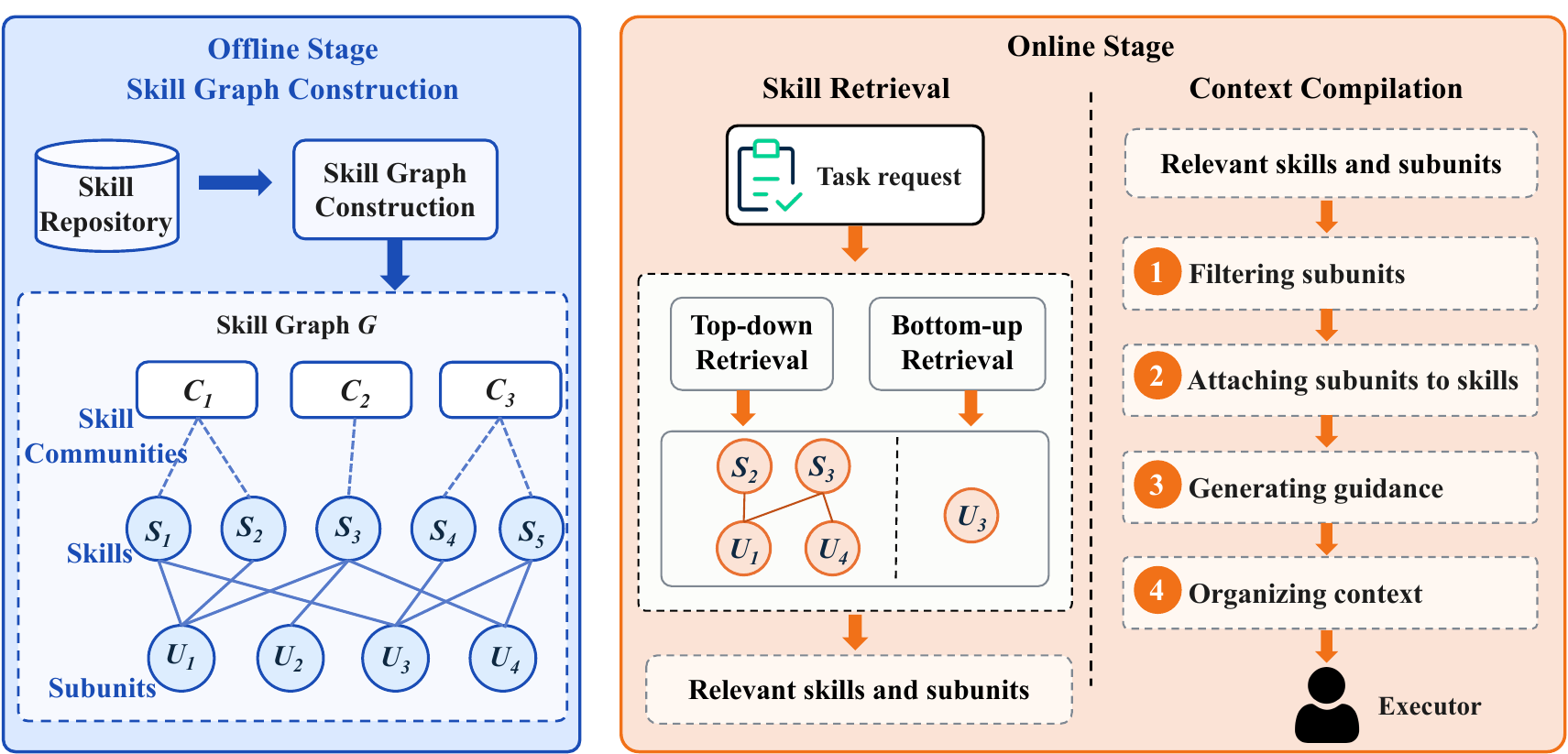}
    \caption{
    Overview of \ourMethod{}.
    }
    \label{fig:SkillRAE-overview}
\end{figure}

Given a task request \(q\), a task environment, and a repository of reusable skills, the RAE systems often first retrieve relevant skill evidence, then compile it into a task-specific context, and finally invoke an executor to complete $q$.
Following this general framework, we propose \ourMethod{}, a two-stage approach that consists of an offline stage and an online stage, where the former one builds an index while the latter one performs skill retrieval and context compilation, as depicted in Figure~\ref{fig:SkillRAE-overview}.

Specifically, in the offline stage, we construct a multi-level skill graph $\mathcal{G}$ over skill communities, skills, and reusable subunits.
This graph preserves skills as executor-compatible anchors while exposing fine-grained subunit evidence that can later support retrieval and context compilation.

In the online stage, we first perform skill retrieval over the constructed graph above by combining evidence from skill communities and subunits, which are retrieved in top-down and bottom-up manners, respectively.
It then compiles the retrieved skills, selected subunit evidence, rescued subunits from non-selected source skills, and task-output constraints into a task-specific context $\mathcal{C}(q)$. 

In the following, we introduce the details of both offline and online stages.

\subsection{Offline Stage}
\label{sec:m1-representation}

In the offline stage, to capture the intricate relationships between different skills, we construct a skill graph, which is a multi-level graph representing the skill repository at three evidence levels, i.e., skill communities, skills, and subunits. 
Formally, we denote the repository representation as
\begin{equation}
    \mathcal{G}=(C,S,U,E,m),
\end{equation}
where \(C\) is the set of skill communities, \(S\) is the set of skill nodes, \(U\) is the set of normalized subunit nodes,
\(E\subseteq S\times U\) denotes the edges between skills and subunits, and \(m:S\rightarrow C\) assigns each skill to one skill community.
An extraction edge \((s,u)\in E\) means that subunit \(u\) is extracted from skill \(s\), while \(m(s)\) records the skill community assignment used later during retrieval.

During the construction, we convert the original skills into skill nodes and candidate subunits.
Candidate subunits are extracted by deterministic rules over procedural lines, element references, and constraint-like usage statements; they are then normalized, filtered by length, deduplicated by exact match, and linked to their source skills through extraction edges.
Additionally, we also store skill descriptions, normalized subunit nodes, subunit embeddings, and skill--subunit extraction edges.

To form skill communities, we derive a compact textual representation for each skill from high-IDF extracted subunits, embed these skill representations, and cluster the embedded skills; the resulting hard clusters serve as the skill communities, with one community assignment per skill.

In the high-level view, the skill communities can be regarded as collections of related skills, skills as executor-compatible procedures, and subunits as local procedures (e.g., file conventions and constraints).
As a result, the skill communities support matching over related skills, skill nodes preserve the procedures consumed by the downstream executor, and subunits provide local procedures which are not visible from skill-level representations alone.
Besides, the many-to-many extraction edges \(E\) allow normalized subunits to be shared across skills, enabling bottom-up evidence aggregation during retrieval and allowing us to incorporate task-specific information from both selected skills and rescued evidence.

\subsection{Online Stage}
\label{sec:m2-retrieval}

In the online stage, whenever the agent receives a task request \(q\), it first retrieves the relevant skills from the skill graph, then compiles the relevant skills into a compact, grounded, and immediately usable form, and finally invoke the executor to complete $q$.
Next, we introduce the details of skill retrieval and context compilation sequentially.

\subsubsection{Skill Retrieval}
\label{sec:skillRetrieval}

Our skill retrieval method uses two signals: a top-down signal from skill communities and a bottom-up signal from subunits.
The output is a selected skill set \(K(q)\), selected-skill highlights \(H(q)\), and non-selected rescue candidate records \(R(q)\) for context compilation.

\paragraph{Top-down retrieval.}
It matches the task request \(q\) to the skill communities in the multi-level skill graph, where each community is represented by a deterministic text constructed from its label and representative skills. 
Communities are scored by embedding similarity between \(q\) and the community text; the highest-scoring communities define the community-supported skill set \(\mathcal{B}(q)\), which later enters skill scoring.
%

\paragraph{Bottom-up retrieval.}
It matches $q$ to subunits and projects their evidence back to source skills. 
Let \(\ell_1(s,q)\) denote normalized similarity between \(q\) and the description of skill \(s\), let \(p(s,q)\) denote a normalized skill name score, and let \(\ell_0(s,q)\) denote normalized subunit evidence for skill \(s\). 
For the retained top subunits \(\mathcal{T}_{N}(q)\), the projection is
\begin{equation}
\ell_0(s,q)=\operatorname{norm}_{S}\!\left(\sum_{u \in \mathcal{T}_{N}(q)} \mathbf{1}\{(s,u)\in E\}\,\frac{\sigma(q,u)}{\deg(u)}\right),
\label{eq:l0-projection}
\end{equation}
where \(\sigma(q,u)\) is similarity between task $q$ and subunit $u$, $deg(u)$, is the number of skills from which subunit $u$ is extracted, and $\operatorname{norm}_{S}$ normalizes scores over skills. 
A subunit $u$ contributes to a skill $s$ only through an edge between them in the multi-level skill graph.

\paragraph{Skill selection.}
The final skill score combines skill-level similarity, bottom-up subunit evidence, the skill name score, and the top-down community-level weighting score:
\begin{equation}
\operatorname{score}(s,q)
= \left(\alpha \ell_1(s,q)+\beta \ell_0(s,q)+\gamma p(s,q)\right)
\left(1+\lambda\mathbf{1}\{s\in \mathcal{B}(q)\}\right),
\label{eq:retrieval-score}
\end{equation}
where \(\mathcal{B}(q)\) is the set of skills assigned to the highest-scoring skill communities matched by the top-down retrieval direction.
The selected skill set is
\begin{equation}
K(q)=\operatorname{TopK}_{s\in S}\operatorname{score}(s,q).
\label{eq:topk}
\end{equation}
Skills remain the units consumed by the executor, while subunits provide evidence for scoring and later context compilation.

\paragraph{Evidence export.}
For each selected skill \(s\in K(q)\), the retriever records a bounded set \(H(s,q)\) of task-relevant subunit highlights extracted from \(s\). 
These highlights come from the retained subunits that contribute to \(\ell_0(s,q)\). 
The retriever ranks them by contribution and task similarity, removes duplicate subunits, and keeps only a small set for each skill. 
They do not replace the selected skill itself; they make local procedures explicit for context compilation. 
Besides, the retriever keeps non-selected candidate records \(R(q)\), which are used later for subunit filtering in the compilation stage.
This design makes the retrieval stage both executor-compatible and compiler-ready: the output remains a selected skill set, while the compiler receives graph-grounded subunit evidence needed to form a task-specific context.

\subsubsection{Context Compilation}
\label{sec:m3-compilation}

Given the selected skills \(K(q)\), subunit evidence \(H(q)\), non-selected rescue candidates \(R(q)\), and task-output contract evidence \(\mathcal{O}(q)\), we aim to compile them into a compact, grounded, and immediately usable form.
This process does not change a source skill, planner, or executor; instead, it generates an effective context that includes task-specific guidance on how to use the selected skills and subunit cues, together with local evidence, for the executor.
As a result, the compilation process changes what the agent sees before execution, rather than the runtime orchestration.

To achieve the above goal, we develop an effective compilation method, which consists of four steps:  filtering subunits, attaching subunits to skills, generating task-specific guidance, and organizing context, introduced as follows.

\paragraph{Step 1: filtering subunits.}
This step handles the case where a skill is not selected into \(K(q)\), but contains a local subunit that is highly relevant to $q$.
Because retrieval selects whole skills as the primary context unit, excluding a source skill from \(K(q)\) does not imply that all subunits extracted from that skill are useless; it only means that the skill was not selected under the aggregate skill score. 
To filter subunits, we scan a bounded pool of non-selected source skills and select a small intermediate set \(Z(q)\) of subunits, subject to relevance, non-redundancy, source skill alignment, and budget constraints. 
Each filtered subunit keeps the identifier of the non-selected source skill it was extracted from, but \(Z(q)\) is not exposed directly as a standalone context.

\paragraph{Step 2: attaching subunits to skills.}
The second step attaches each filtered subunit to a selected skill.
Specifically, for each \(u\in Z(q)\), we choose the selected skill with the highest compatibility:
\begin{equation}
s^{\star}(u)=
\arg\max_{s\in K(q)} \operatorname{Aff}(u,s\mid q),
\label{eq:affiliation}
\end{equation}

where \(\operatorname{Aff}(u,s\mid q)\) is a deterministic compatibility score over a small feature vector:
\begin{equation}
\operatorname{Aff}(u,s\mid q)
=
\boldsymbol{\omega}^{\top}
\big[
q_{\mathrm{rel}}(u,q),\,
f_{\mathrm{skill}}(u,s,H(s,q)),\,
r_{\mathrm{sel}}(s,q),\,
g_{\mathrm{same}}(u,s),\,
g_{\mathrm{active}}(u,s,K(q))
\big].
\label{eq:affscore}
\end{equation}
Here \(\boldsymbol{\omega}\) is a fixed weight vector, \(q_{\mathrm{rel}}\) measures lexical relevance between subunit and task, \(f_{\mathrm{skill}}\) measures the best match between \(u\) and the selected skill \(s\) or its highlighted subunits, \(r_{\mathrm{sel}}\) is the normalized retrieval confidence of \(s\), and \(g_{\mathrm{same}}\) and \(g_{\mathrm{active}}\) encode skill community alignment. Implementation details of these features are provided in Appendix~\ref{app:affiliate-attachment}.

The attachment step turns rescued subunits into selected-skill-local cues.
For each rescued subunit \(u\), the compiler records the selected skill \(s^{\star}(u)\) to which it is attached, together with the original non-selected source skill \(p\) from which \(u\) was extracted.
We denote the resulting pool of attached candidate cues as \(\mathcal{A}(q)\).
At this point, \(\mathcal{A}(q)\) is not yet the set rendered in the compiled context; it is the candidate pool passed to the guidance-generation step.
This attachment step makes rescue compatible with the selected-skill interface: rescued evidence is associated with an already selected skill, not exposed as a new skill or independent execution unit.

\paragraph{Step 3: Generating task-specific guidance.}
This step selects which attached cues in \(\mathcal{A}(q)\) should be rendered as task-specific guidance.
The compiler keeps cues that are concrete, task-aligned, non-redundant with already selected evidence, and useful under the context budget.
This produces \(\mathcal{A}_{\mathrm{final}}(q)\), the set of affiliated cues that will be rendered in the compiled context.
Low-affinity or overly generic cues are excluded even if they were recovered during rescue, preventing the compiled context from becoming a dump of all available subunit evidence.

\paragraph{Step 4: organizing context.}
The last step organizes the selected evidence and task-specific guidance into the compiled context.
\begin{equation}
\mathcal{C}(q)=
\operatorname{Compile}_{B}\!\left(
q,\,
K(q),\,
H(q),\,
\mathcal{A}_{\mathrm{final}}(q),\,
\mathcal{O}(q)
\right),
\label{eq:context-compile}
\end{equation}

where \(B\) is the context budget. 
The compiled context is organized around the task request, selected skills and subunits, output contract, and affiliated subunit cues.
The compiled context is delivered to the downstream executor as advisory context before execution.

Overall, in the online stage, we not only retrieve the skills that are highly relevant to $q$, but also compile them into the context with task-specific guidance on how to apply them to complete $q$. 
%

\section{Experiments}
\label{sec:exp}
\providecommand{\pending}{\textemdash}
\providecommand{\best}[1]{\textbf{#1}}
\providecommand{\pp}{\,pp}

We evaluate \ourMethod{} to test whether context compilation improves downstream skill usage.
The experiments compare overall performance, isolate the contribution of compilation and graph evidence, and include a targeted cross-backbone check over a second agent/model backbone.

\subsection{Setup}
\label{sec:exp-setup}

\textbf{Datasets.}
We evaluate on two complementary skill agent benchmarks. 
\textsc{SkillsBench} is the primary execution-centric benchmark because each task defines a deterministic verifier and produces a scalar reward, allowing us to measure downstream task completion rather than retrieval accuracy alone \citep{xli2026skillsbench}. 
\textsc{AgentSkillOS} provides a complementary skill quality benchmark for ecosystem-scale skill use \citep{hli2026agentskillos}. 
Table~\ref{tab:dataset-statistics} summarizes the benchmark and graph statistics used in our experiments.

\begin{table}[t]
\centering
\small
\caption{Benchmark and graph statistics. \#L2, \#L0, and \#Edges denote the number of skill communities, subunits, and extracted subunits.}
\label{tab:dataset-statistics}
\begin{tabular}{lcccccc}
\toprule
Benchmark & \#Tasks & Metric & \#Skills & \#L2 & \#L0 & \#Edges \\
\midrule
SkillsBench & 87 & Reward Mean & 207 & 14 & 4,834 & 5,255 \\
AgentSkillOS & 30 & Score & 200 & 14 & 718 & 918 \\
\bottomrule
\end{tabular}
\end{table}

\textbf{Evaluation and metrics.}
Following existing works~\citep{xli2026skillsbench,hli2026agentskillos}, we evaluate each method through end-to-end agent execution.
For each method, the selected skills, along with any compiled guidance provided, are sent to the same downstream agent executor, which executes the task and produces the required output artifacts.
The resulting artifacts are evaluated against the task-specific ground truth or expected constraints defined by each benchmark.
For SkillsBench, we report the mean reward across 87 tasks, where each task is evaluated by a deterministic verifier that assigns a scalar reward based on artifact-level correctness.
For AgentSkillOS, we report the aggregate score over 30 tasks, where weighted objective checks are mapped to a normalized score from 0 to 100.
This evaluation measures the extent to which the retrieved skills and compiled guidance can be operationalized by the agent to complete the target tasks, rather than only assessing retrieval coverage.

\textbf{Baseline methods.}
We compare SkillRAE with five baselines. 
\emph{Benchmark-native curated skills} (from SkillsBench~\cite{xli2026skillsbench} and AgnetSkillOS~\cite{hli2026agentskillos}) denotes the curated skills that are manually specified per-task in each dataset.
\emph{Vanilla retrieval} embeds each skill using its frontmatter name and description, ranks skills by embedding similarity to the task request, and exposes the selected skills directly to the executor. 
\emph{LLM-based retrieval} first uses the same embedding retriever to form a candidate pool, then asks an LLM selector to choose the final skills from candidate skill IDs, names, and descriptions; the selected skills are then exposed directly to the executor.
\emph{AgentSkillOS} \citep{hli2026agentskillos} is evaluated in its native setting, using its built-in skill organization and orchestration pipeline; we do not add context compilation because doing so would change the execution-control surface being evaluated. 
\emph{SkillRouter} \citep{zheng2026skillrouter} is evaluated as a full-text skill-routing baseline: it produces a ranked skill set, which we materialize for the same downstream executor. 
%
%

\textbf{Implementation.}
We build the multi-level skill graph offline from source \texttt{SKILL.md} files.
The graph builder extracts normalized procedural, element, and constraint subunits from each skill, deduplicates them globally, and connects each retained subunit back to its source skills. 
We embed subunits with \texttt{BAAI/bge-small-en}~\cite{xiao2023cpack} using normalized SentenceTransformer embeddings \citep{xiao2023cpack,reimers2019sentencebert}.
Skill descriptions are embedded with the same model at retrieval time.
Skill communities are constructed by embedding skill representations and applying KMeans clustering.
At retrieval time, the system retrieves a bounded set of skills, exports a bounded set of highlighted subunits for each selected skill, and applies rescue-aware compilation over a bounded non-selected frontier.
Exact retrieval budgets, clustering settings, rescue thresholds, and final-packet budget are provided in Appendix~\ref{app:implementation-details}.
we use Codex CLI with GPT-5.2 \citep{openai2026codexcli,openai2025gpt52} as the downstream agent/model backbone. 

\subsection{Overall Performance}
\label{sec:overall-performance}

We first compare \ourMethod{} with skill use and retrieval baselines under the same downstream execution setting.  Table~\ref{tab:main-results} reports the main comparison results.  The table uses one primary metric per benchmark: SkillsBench reward mean and AgentSkillOS aggregate score.

\begin{table}[t]
\centering
\small
\caption{Overall downstream performance.  SkillsBench reports reward mean over 87 tasks.  AgentSkillOS reports aggregate score under the score-based evaluation protocol.}
\label{tab:main-results}
\begin{tabular}{@{}lcc@{}}
\toprule
Method & SkillsBench Reward Mean (\%) $\uparrow$ & AgentSkillOS Score (\%) $\uparrow$ \\
\midrule
Benchmark-native Curated Skills & 26.20 & 78.45 \\
AgentSkillOS & 17.24 & 83.50 \\
Vanilla Retrieval & 19.44 & 74.85 \\
LLM-based Retrieval & 16.34 & 82.06 \\
SkillRouter & 22.04 & 82.30 \\
\ourMethod{} (Ours) & \best{29.26} & \best{84.59} \\
\bottomrule
\end{tabular}
\end{table}

\ourMethod{} achieves the best overall performance on both benchmarks. 
On SkillsBench, it improves over benchmark-native curated skills by 3.06 percentage points, over SkillRouter by 7.22 percentage points, and over vanilla retrieval by 9.82 percentage points. 
On AgentSkillOS, it also obtains the highest aggregate score, improving over the native AgentSkillOS baseline by 1.09 percentage points and over SkillRouter by 2.29 percentage points.
The strongest baselines are informative. 
Benchmark-native curated skills provide a strong reference because the relevant skills are manually specified or benchmark-defined. 
SkillRouter is competitive because full-text skill routing improves skill selection, while AgentSkillOS performs strongly on its native benchmark because it uses an orchestration-oriented execution pipeline. 
However, these methods still expose selected skills or orchestrated skill use without explicitly compiling retrieved skill evidence into a compact, task-specific execution context. 
The fact that \ourMethod{} improves even over benchmark-native curated skills suggests that task success depends not only on selecting relevant skills, but also on how the selected skill evidence is organized for the executor.

These results indicates that effective retrieval-augmented execution requires both strong skill retrieval and context compilation. 
\ourMethod{} improves downstream task completion by converting retrieved skills, subunit evidence, and task constraints into a compact context.

\subsection{Detailed Analysis of \ourMethod{}}
\label{sec:component-ablations}

\textbf{Ablation study.}
To understand the effectiveness of different components in \ourMethod{}, we conduct ablation studies and report the results in Table~\ref{tab:component-ablation}.
The first two ablation variants remove the Bottom-up Retrieval and Top-down Retrieval modules from the Skill Retrieval stage, respectively, while the variant \texttt{w/o Context Compilation} removes the Context Compilation stage.
Among all variants, removing Top-down Retrieval causes the largest performance degradation, reducing the score from 29.26\% to 16.61\% with Codex CLI + GPT-5.2 \citep{openai2026codexcli,openai2025gpt52}.
This confirms the value of skill communities, which support coarse-grained matching over related skills before the agent selects and applies specific skill units.
The drops observed when removing Bottom-up Retrieval further show that fine-grained matching remains useful, while the degradation under \texttt{w/o Context Compilation} indicates that compiling retrieved skills into a task-specified guidance is important for effective skill application.

\begin{table}[h]
\centering
\small
\caption{Component ablations across agent/model backbones on SkillsBench. Each row removes one method-level component from the full graph-native context compilation method. All entries report reward mean (\%).}

\label{tab:component-ablation}
\begin{tabular}{@{}lcc@{}}
\toprule
Variant & Codex CLI + GPT-5.2 & Gemini CLI + Gemini 3 Flash \\
\midrule
\ourMethod{} Full & \best{29.26} & \best{28.85} \\
-- \texttt{w/o Bottom-up Retrieval} & 23.43 & 25.57 \\
-- \texttt{w/o Top-down Retrieval} & 16.61 & 20.26 \\
-- \texttt{w/o Context Compilation} & 22.59 & 27.80 \\
\bottomrule
\end{tabular}
\end{table}

\textbf{Performance under different agent backbones.}
Table~\ref{tab:component-ablation} further demonstrates the performance under different agent backbones.
Table~\ref{tab:component-ablation} also evaluates the same ablation design under two agent/model backbones: Codex CLI with GPT-5.2 \citep{openai2026codexcli,openai2025gpt52} and Gemini CLI with Gemini 3 Flash \citep{google2026geminicli,google2025gemini3flashcli}.
The full method achieves similar rewards under the two backbone, 29.26\% and 28.85\%, indicating that the overall pipeline is not tied to a single agent backbone.
All three ablations reduce performance under both backbones, indicating consistent component effects.
At the same time, the Gemini-side drops are smaller for some components, especially Context Compilation, suggesting that the downstream agent/model backbone can partially mediate the effect size of missing context-formation signals.
Nevertheless, Top-down Retrieval remains the most important module under both backbone, with the largest absolute degradation in both columns.
%




\subsection{Context Compilation on Compatible External Backbones}
\label{sec:context-backbones}

We next evaluate lite context compilation on compatible external retrieval backbones. A compatible backbone outputs a flat selected skill set that is consumed by the same executor. This criterion includes vanilla retrieval and LLM-based retrieval, but excludes SkillRouter and AgentSkillOS: SkillRouter is kept as a full-text routing baseline, and AgentSkillOS is kept as an orchestration baseline. Table~\ref{tab:context-backbones} therefore tests the lite compilation interface only where the intervention is well-defined.

The context compilation results show that selected skills become more useful when they are surfaced through a compact task/output-aware exposure layer.  This effect appears for both vanilla retrieval and LLM-based retrieval, where the compiler has access only to selected skills and task/output contracts.  
The \ourMethod{} pair tests the stronger version of the same idea: given our graph-native retrieval backbone, affiliate-aware compilation further improves reward from 22.59\% to 29.26\%.  
This separates the general value of context exposure from the additional value of graph-derived subunit export and affiliate-aware rescue.

\begin{table}[h]
\centering
\small
\caption{Effect of context compilation across retrieval backbones.  External backbones use lite context compilation without graph-derived affiliation.  \ourMethod{} Full uses graph-native affiliate-aware context compilation.}
\label{tab:context-backbones}
\begin{tabular}{@{}llcc@{}}
\toprule
Retrieval Backbone & Setting & SkillsBench Reward Mean (\%) $\uparrow$ & Gain (\%) $\uparrow$ \\
\midrule
Vanilla retrieval & No compilation & 19.44 & -- \\
Vanilla retrieval & + Context compilation & 21.71 & +2.27 \\
\midrule
LLM-based Retrieval & No compilation & 16.34 & -- \\
LLM-based Retrieval & + Context compilation & 23.36 & +7.02 \\
\midrule
\ourMethod{} & Retrieval only & 22.59 & -- \\
\ourMethod{} & + Context compilation & 29.26 & +6.67 \\
\bottomrule
\end{tabular}
\end{table}

\subsection{Summary of findings.}

The experiments support three aspects of the method. First, \ourMethod{} improves over existing skill use, retrieval, routing, and orchestration baselines in the main benchmark comparison.  
Second, the ablations show that the gain depends on both context compilation and graph-derived evidence sources, including subunit evidence and L2 skill communities.
Third, the context compilation improves compatible selected-skill exposure methods, while \ourMethod{} further strengthens this exposure layer with graph-native affiliate-aware evidence.  
Together, these findings support the paper's thesis that retrieval-augmented execution requires context compilation beyond skill selection: retrieved skills and subunits must be organized into a compact, grounded, and task-specific context for execution.

\section{Limitations}
\label{sec:limitation}

\ourMethod{} is most applicable to skill repositories with explicit procedural text, file/output conventions, and constraints; it may be less effective when key dependencies are hidden in opaque tools, undocumented code, or runtime state.
As an advisory compiler rather than a planner or controller, \ourMethod{} does not guarantee downstream context use or recovery from execution-time failures; future work can evaluate it on larger repositories, more executor backbones, and broader skill-driven benchmarks as they become available.

\section{Conclusion}
\label{sec:conclusion}

In this paper, we study the skill-based context compilation for retrieval-augmented execution (RAE).
We propose \ourMethod{}, a two-stage RAE approach that constructs a multi-level skill graph in the offline stage, and then in the online stage retrieves relevant skills and subunits and further compiles them into a compact, grounded, and task-specific context with task-specified guidance.
%
%
Experiments on SkillsBench and AgentSkillOS demonstrate that \ourMethod{} improves downstream execution performance over baseline methods. 
In future work, we will test \ourMethod{} on more skill benchmarks and also consider more real-world constraints such as limited token budget.

\clearpage
\bibliographystyle{unsrtnat}
\bibliography{references}

\newpage
\appendix
\section{Appendix}

\subsection{Implementation Details}
\label{app:implementation-details}
The subunit extractor is deterministic. It collects three types of support evidence from each source \texttt{SKILL.md}: procedural lines, element-like strings such as file names, command-line patterns, and library names, and short constraint-like lines matched by requirement keywords.
It normalizes text, removes duplicate subunits, filters subunits by token length, and creates skill--subunit extraction edges.
Skill descriptions used for skill-level representations are extracted from the YAML frontmatter of each source \texttt{SKILL.md}.

For L2 community construction, KMeans uses \texttt{random\_state}=42 and \texttt{n\_init}=10, with the number of clusters set to $k=\lfloor\sqrt{|S|}\rfloor$. Rescue selection uses a parent-score threshold of 0.35, a subunit score threshold of 0.12, a global rescue cap of 3, a per-parent cap of 1, and a token-Jaccard redundancy threshold of 0.6. The rescue candidate pool is bounded by
\[
\min(|\mathrm{ranked}|,\max(10,4k)).
\]
The context budget is 384 tokens.

For retrieval baselines, we use top-$k=5$ selected skills. The LLM-based retrieval baseline first forms a 32-skill candidate pool with embedding retrieval before LLM selection.

\subsection{Implementation Details of Affiliate Attachment}
\label{app:affiliate-attachment}

This appendix gives the implementation details behind the affiliate attachment score used in Section~\ref{sec:m3-compilation}. Affiliate attachment is a deterministic post-retrieval context attachment layer: it does not change the selected skill set, does not perform random walk or graph diffusion, and does not introduce new executable skills. Given a rescued subunit \(u\) and a candidate selected skill \(s\), the implementation scores the attachment by a fixed weighted combination of five features:
\[
\begin{aligned}
\operatorname{AffScore}_{\mathrm{impl}}(u,s\mid q)
={}&
0.15\,q_{\mathrm{rel}}(u,q)
+
0.45\,f_{\mathrm{skill}}(u,s,H(s,q)) \\
&+
0.10\,r_{\mathrm{sel}}(s,q)
+
0.15\,g_{\mathrm{same}}(u,s)
+
0.15\,g_{\mathrm{active}}(u,s,K(q)).
\end{aligned}
\]
Here \(q_{\mathrm{rel}}\) is token-overlap relevance between the rescued subunit text and the task request. \(f_{\mathrm{skill}}\) is the maximum of the overlap between the rescued subunit and the selected skill profile, and the best overlap between the rescued subunit and the selected skill's highlighted subunits. \(r_{\mathrm{sel}}\) is the min--max normalized retrieval confidence of the selected skill. \(g_{\mathrm{same}}\) indicates whether the rescued subunit's source skill and the candidate selected skill belong to the same L2 capability group. \(g_{\mathrm{active}}\) indicates whether the rescued subunit is consistent with the active L2 capability groups induced by the selected skills.

For each rescued subunit, we score all selected skills and attach the subunit to the selected skill with the highest score. The implementation then applies a small exclusivity bonus when the best selected skill is clearly separated from the second-best selected skill, and applies a penalty when the rescued subunit falls outside the active L2 capability groups. The resulting attachment is accepted only if it passes the affiliation threshold and later compact-packet gating. Compact gating controls whether an accepted affiliated cue is rendered in the final packet; it requires the cue to be concrete, task-aligned, non-redundant, and able to fit within the optional-cue token budget.

\subsection{Reproducibility Details}
\label{app:reproducibility}

We use the benchmark-provided task definitions and fixed task lists for all reported experiments. SkillsBench results are reported over 87 evaluated tasks, and AgentSkillOS results are reported over 30 evaluated tasks under the benchmark score-based evaluation protocol. We do not remove tasks after observing model outputs. SkillsBench uses its deterministic verifier reward as the task outcome, and we report the macro-average reward over the fixed task set. AgentSkillOS results use the aggregate score produced by its native benchmark evaluation protocol.

\paragraph{Data and skill pools.}
For benchmark-native curated skills, we use the curated reference condition defined by each benchmark: per-task oracle skill folders in SkillsBench and the benchmark-defined native skill group in AgentSkillOS. For retrieval-based methods, the skill pool is loaded from the benchmark skill repositories, where each skill is represented by a folder containing a \texttt{SKILL.md} file and optional auxiliary resources. The same task environment, skill-pool exposure protocol, runtime budget, downstream executor, and aggregation rule are used for comparable methods within each benchmark.

\paragraph{Graph construction.}
The multi-level skill graph is constructed offline from source \texttt{SKILL.md} files. The graph construction pipeline extracts normalized procedural, element-like, and constraint-like subunits, deduplicates them globally, and connects each retained subunit back to its source skills. Subunits and canonical skill descriptions are embedded with \texttt{BAAI/bge-small-en} using normalized SentenceTransformer embeddings. L2 skill communities are obtained by embedding skill representations and applying KMeans with $k=\lfloor\sqrt{|S|}\rfloor$, \texttt{random\_state=42}, and \texttt{n\_init=10}. The resulting artifacts include skill nodes, subunit nodes, skill--subunit edges, canonical skill representations, skill--community assignments, subunit identifiers, and subunit embeddings.

\paragraph{Retrieval and compilation settings.}
Unless otherwise stated, retrieval uses top-$k=5$ selected skills. The retriever considers the top 30 task-matched subunits, keeps the top 2 L2 communities as the community-level signal, and exports at most three highlighted subunits per selected skill. Rescue-aware compilation uses a parent-score threshold of 0.35, a subunit score threshold of 0.12, a global rescue cap of 3, a per-parent cap of 1, and token-Jaccard redundancy filtering. Affiliate attachment uses fixed compatibility weights and accepts cues only after affiliation and compact-packet gating. The context budget is 384 tokens.

\paragraph{Baselines and compilation protocol.}
All retrieval-backend methods are implemented through \texttt{experiments/retrieval\_tasks\_backend/run\_retrieval\_tasks\_backend.sh} and its associated mirror-preparation and aggregation scripts. Vanilla top-$k$ retrieval embeds each skill using its frontmatter name and description, ranks skills by similarity to the task request, and exposes the selected skills to the unchanged executor. LLM-based retrieval first builds an embedding-retrieval candidate pool and then asks an LLM selector to choose the final skill IDs from candidate names and descriptions. Flat retrieval baselines are additionally evaluated with a lite context protocol, which renders selected skills, task/output-contract information, and a short execution checklist into an instruction overlay without using graph-derived subunit export, rescue, or affiliate attachment. Full SkillRAE uses the graph-native context protocol implemented in the same retrieval backend: it renders selected skills, selected-skill subunit highlights, output-contract information, and affiliated rescue cues into the per-task \texttt{READ\_FIRST.md} / \texttt{COORDINATOR\_PACKET.json} payload delivered to the same executor. SkillRouter and AgentSkillOS are kept in their native forms rather than being augmented with our compiler.

\paragraph{Agent/model substrates and execution.}
The main comparisons use Codex CLI with GPT-5.2 \citep{openai2026codexcli,openai2025gpt52} as the downstream agent/model substrate. Component ablations additionally instantiate the same SkillRAE variants with Gemini CLI and Gemini 3 Flash. If decoding parameters are not explicitly configured by the CLI backend, we use provider defaults. The retrieval backend stages task mirrors, materializes the selected skills and context overlays, launches the benchmark executor, collects verifier outputs, and writes row-level summaries. Command-level timeouts and failure buckets are recorded in the run summaries; verifier failures after task execution are treated as benchmark failures rather than removed from the task set.

\paragraph{Logging and verification artifacts.}
Each run writes per-task summaries, selected-skill metadata, context artifacts, executor logs, verifier outputs, and run-level aggregation files under the corresponding output directory. These artifacts are sufficient to verify the reported SkillsBench reward means by recomputing the mean verifier reward over the fixed task list. For AgentSkillOS, we report the aggregate score from the benchmark's native evaluation protocol.

\subsection{Compute Resources}
\label{app:compute}

All experiments were orchestrated on a Linux server running Ubuntu 24.04 LTS
with Linux kernel 6.8.0. The worker machine has two Intel Xeon Gold 6330 CPUs
with 56 physical cores and 112 logical threads in total, 1.0 TiB of RAM,
Docker 29.3.0, Docker Compose v2.29.7, and Python 3.12.9. The machine also
has eight NVIDIA RTX A5000 GPUs, but the inspected retrieval and graph
construction code paths use CPU execution for local embedding and retrieval
operations.

The local worker handles graph preparation, retrieval, context compilation,
task-mirror construction, containerized task execution orchestration, verifier
execution, logging, and result aggregation. Hosted LLM inference is performed
through the external Codex CLI \citep{openai2026codexcli} and Gemini CLI \citep{google2026geminicli} backends. Runs are launched by
the benchmark runners with configurable parallelism; the audited retrieval
backend uses one worker by default, while some archived launch wrappers use up
to four workers.

\begin{table}[t]
\centering
\small
\caption{Compute resources for the reported experiments. Wall-clock time is measured from available run logs and artifact timestamps when recoverable.}
\label{tab:compute-resources}
\begin{tabular}{@{}p{0.24\linewidth}p{0.25\linewidth}p{0.43\linewidth}@{}}
\toprule
Experiment & Wall-clock & Notes \\
\midrule
Offline graph construction
& Not separately isolated
& Graph construction and embedding preparation are lightweight relative to agent execution and were not logged as a standalone timed phase. \\

SkillsBench main runs
& 3.4--5.0 hours per 87-task large run
& End-to-end time includes retrieval, context construction, containerized execution, verifier collection, and aggregation. \\

Component ablations
& 0.3--2.8 hours per 8--10 task batch
& Time varies with task subset, backend, and execution failures. \\

Context comparison
& Not separately isolated
& Timing was not cleanly separable from the corresponding retrieval-backbone runs in the available artifacts. \\

AgentSkillOS runs
& Not separately isolated
& A standalone wall-clock trace was not recovered in the bounded audit. \\

Gemini cross-substrate check
& Not separately isolated
& Gemini execution is supported by the runner configuration, but dedicated timing artifacts were not clearly separated from the sampled logs. \\
\bottomrule
\end{tabular}
\end{table}

Across the bounded set of sampled run directories, we observed at least 303
planned task-level Harbor invocations and at least 33.66 wall-clock hours of
recorded runs. These totals are conservative because preliminary, failed, and
debugging runs were not exhaustively audited. The reported compute should
therefore be interpreted as the recoverable compute for the table-supporting
runs rather than the full exploratory cost of the project.

\newpage

\end{document}